\documentclass[journal]{IEEEtran}
\usepackage{cite}

\ifCLASSINFOpdf
   \usepackage[pdftex]{graphicx}
\else
\fi
\ifCLASSOPTIONcompsoc
  \usepackage[caption=false,font=normalsize,labelfont=sf,textfont=sf]{subfig}
\else
  \usepackage[caption=false,font=footnotesize]{subfig}
\fi
\usepackage{fixltx2e}
\begin{document}
%
\title{Automatic Channel Network Extraction from Remotely Sensed Images by Singularity Analysis}
%
%
%

\author{Furkan~Isikdogan, Alan~Bovik, Paola~Passalacqua
\thanks{The authors are with the University of Texas at Austin, TX 78112 USA (e-mail: isikdogan@utexas.edu; bovik@ece.utexas.edu; paola@austin.utexas.edu).}
\thanks{Manuscript received April 3, 2015; revised June 15, 2015.}}

\maketitle

\begin{abstract}
Quantitative analysis of channel networks plays an important role in river studies. To provide a quantitative representation of channel networks, we propose a new method that extracts channels from remotely sensed images and estimates their widths. Our fully automated method is based on a recently proposed Multiscale Singularity Index that responds strongly to curvilinear structures but weakly to edges. The algorithm produces a channel map, using a single image where water and non-water pixels have contrast, such as a Landsat near-infrared band image or a water index defined on multiple bands. The proposed method provides a robust alternative to the procedures that are used in remote sensing of fluvial geomorphology and makes classification and analysis of channel networks easier. The source code of the algorithm is available at: http://live.ece.utexas.edu/research/cne/.

\end{abstract}


%
\IEEEpeerreviewmaketitle

\section{Introduction}
%
%
%
%
\IEEEPARstart{A}{} method for completely automatic extraction of channel networks from satellite imagery could greatly facilitate the monitoring of water resources by eliminating the laborious process of manual inspection. Such a method could be used for creating quantitative representations of channel networks, which would be useful in a wide variety of studies. The automatic extraction of channel networks is particularly
challenging in coastal areas due to low topographic gradients, the presence of features such as sediment plumes, and the wide range of scales over which channel features are present. A robust channel extraction method would ease monitoring coastal areas and analyzing deltaic response to anthropogenic and natural forcings over large spatial areas and long temporal intervals.

Several approaches have been suggested to extract curvilinear structures from remotely sensed images, many of them focusing on road network extraction \cite{shackelford2003fully, tupin1998detection, Negri2006, lacoste2005point, poullis2010delineation, valero2010advanced}. Based on the road network extraction work \cite{tupin1998detection, Negri2006}, a method has been proposed to detect rivers as linear structures, imposing constraints on river length, curvature, and confluences for connectivity \cite{Cao2011}. A software tool, \textit{RivWidth} \cite{rivwidth}, has been proposed for calculating river centerlines and widths. \textit{RivWidth} (v0.4) requires the availability of a previously defined binary mask that indicates water and non-water pixels. Although such a mask could be extracted from remotely sensed images by thresholding and using shape-correcting operations, manual cleaning of the mask would often be necessary to separate the true water mask from spurious responses \cite{passalacqua2013geomorphic}. Our new method can estimate channel centerline, width, and orientation, and create a map of a channel network in a purely automatic manner, using only remotely sensed data. To the best of our knowledge, this is the first fully automatic approach that provides these outputs.


\section{Modified Multiscale Singularity Index}
\noindent
The Multiscale Singularity Index \cite{muralidhar2013steerable, muralidhar2013noise} is a useful method for detecting singular curvilinear structures over multiple scales. The algorithm is useful for locating channels in satellite images. However, presence of channels over a wide range of scales creates some artifacts in the Singularity Index response. Our work modifies and extends the Multiscale Singularity Index to address the multi-scale nature of channel networks. The Multiscale Singularity Index algorithm and our modifications are explained briefly in the following sections.

\subsection{The Multiscale Singularity Index}
\noindent
At each pixel, the Multiscale Singularity Index algorithm first estimates the direction $\theta$ orthogonal to the curvilinear mass, using second order derivatives of an input image along evenly spaced directions. Then, it computes the Singularity Index at each scale as:

\begin{equation}
(\psi f)(x,y,\sigma) = \frac{|f_{0,\theta,\sigma}(x,y) f_{2,\theta,\sigma}(x,y)|}{1 + |f_{1,\theta,a\sigma}(x,y)|}
\end{equation}

where $f_{0,\theta,\sigma}(x,y)$, $f_{1,\theta,\sigma}(x,y)$, and $f_{2,\theta,\sigma}(x,y)$ are the zero, first, and second order Gaussian derivatives at scale $\sigma$ and along the direction $\theta(x,y)$. In the denominator, $a$ is a constant with a recommended value 1.7754. This value results in maximum attenuation of the side lobe response of the index \cite{muralidhar2013steerable}. The index $(\psi f)(x,y,\sigma)$ is computed over $N$ scales $\sigma_n = \sigma_1 \sqrt{2}^{(n-1)}$ for $n=1,2,...,N$. The window sizes for the image filters are determined to be $\left \lceil{6 \sigma}\right \rceil$. Since a channel of width larger than the image dimensions cannot be detected by the algorithm, an upper bound for the number of scales $N$ can be determined by having the filter dimensions smaller than the image dimensions $M \times M$ so that $6\sigma_1 \sqrt{2}^{(N-1)} \leq M$:

\begin{equation}
N = \left \lceil{\frac{2 \log{\frac{M}{6 \sigma_1}}}{\log{2}} + 1}\right \rceil
\end{equation}

After computing the Singularity Index at each scale, the algorithm finds the maximum response across all scales at each pixel location.

The Singularity Index retains polarity, which is useful for discriminating between channel and island response, since channels and islands have opposite polarity. By discarding the negative polarity, we remove the island response.

\subsection{Debiasing Input Images}
\noindent
An input image should be debiased before computing the Singularity Index in order to achieve invariance to local DC offset. The Multiscale Singularity Index algorithm debiases an input image by subtracting a large Gaussian filter from the original image, which essentially performs local mean subtraction. This approach works well for a small range of scales. For a large range of scales, however, a large Gaussian filter fails to debias finer scales and results in a loss of detail at fine scales (Fig. \ref{orig_response}). To address this problem, we debias the input image at every scale. Instead of using one large Gaussian filter over all scales, our modified version of the Multiscale Singularity Index uses a Gaussian filter with a standard deviation of $\sigma_n$ at each scale as:
\begin{equation}
I_{\sigma} = I - \mathcal{G}_{\sigma} * I
\end{equation}
where $I_{\sigma}$ is the debiased image at scale $\sigma$, $I$ is the input image, and $\mathcal{G}_{\sigma}$ is the Gaussian filter.

\subsection{Width Estimation}
\noindent
The channel width is estimated by interpolating between the scale that has the highest Singularity Index response $\psi$ and its neighbor scales as follows:
\begin{equation}
w(x,y) = k\frac{\sum\limits_{i=-1}^{+1} \sigma_{m+i} (\psi f)(x,y,\sigma_{m+i})}{\sum\limits_{i=-1}^{+1} (\psi f)(x,y,\sigma_{m+i})}
\end{equation}
where $m = \mathrm{arg\,max}_n (\psi f)(x,y,\sigma_n)$ at spatial coordinate $x, y$ and $k$ is a scalar variable that scales the output.

\subsection{Adaptive Smoothing}
\begin{figure}[!t]
\centering
\subfloat[]{\includegraphics[width=2.95in]{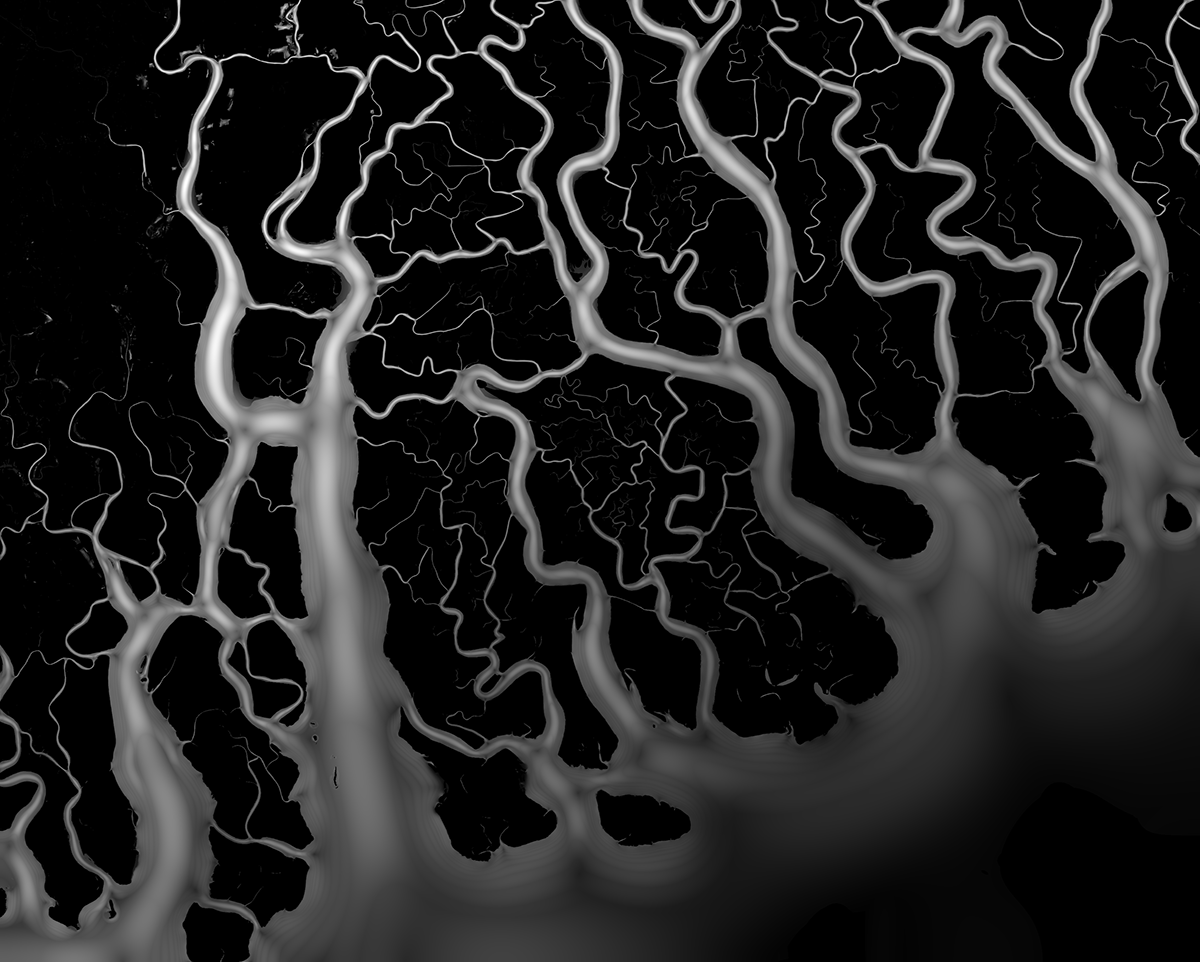}%
\label{orig_response}}
\vspace{-0.05in}
\subfloat[]{\includegraphics[width=2.95in]{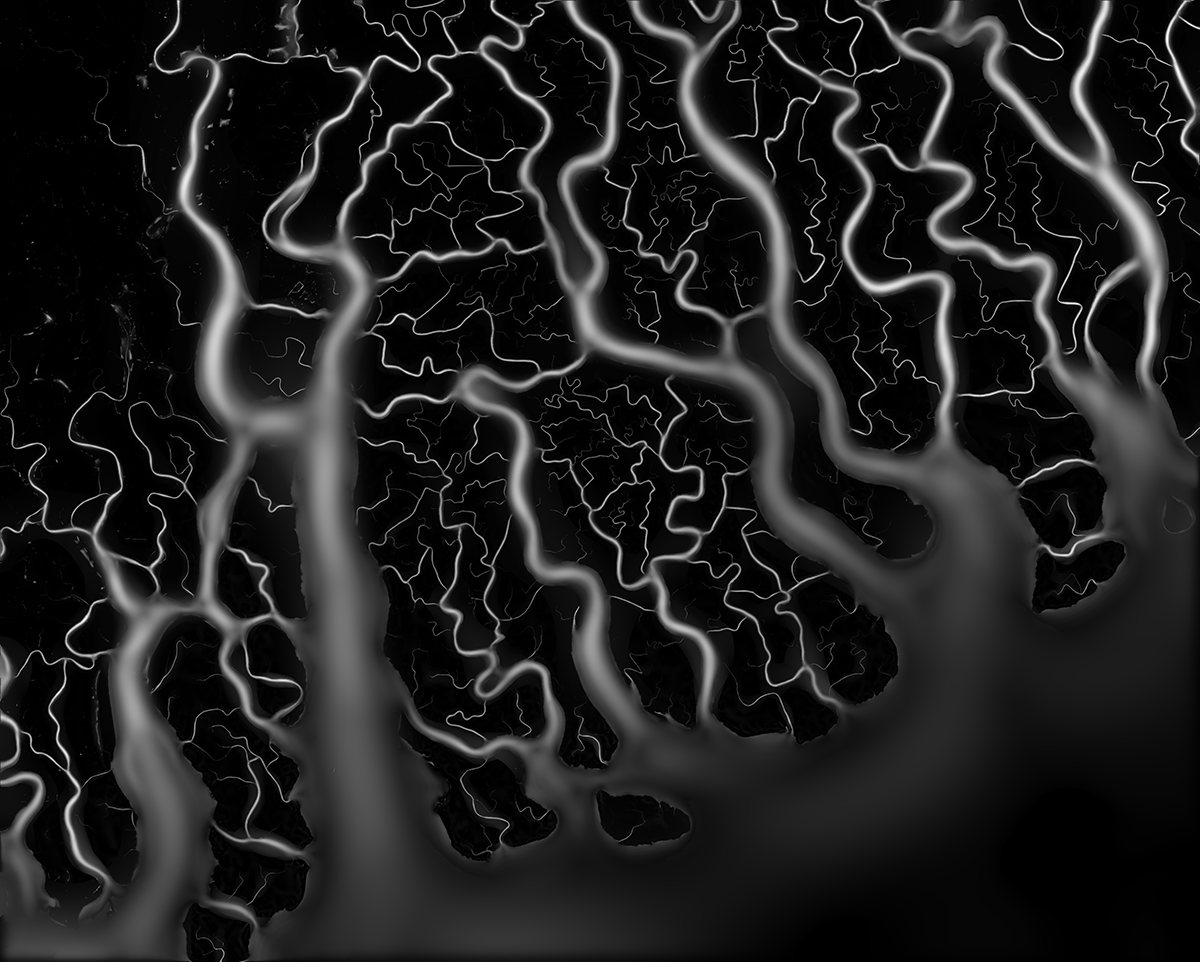}%
\label{mod_response}}
\vspace{-0.05in}
\subfloat[]{\includegraphics[width=1.425in]{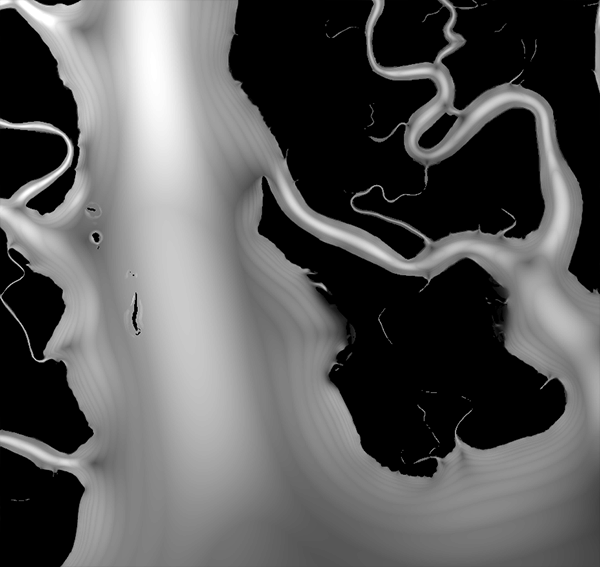}%
\label{orig_response_zoom}}
~
\subfloat[]{\includegraphics[width=1.425in]{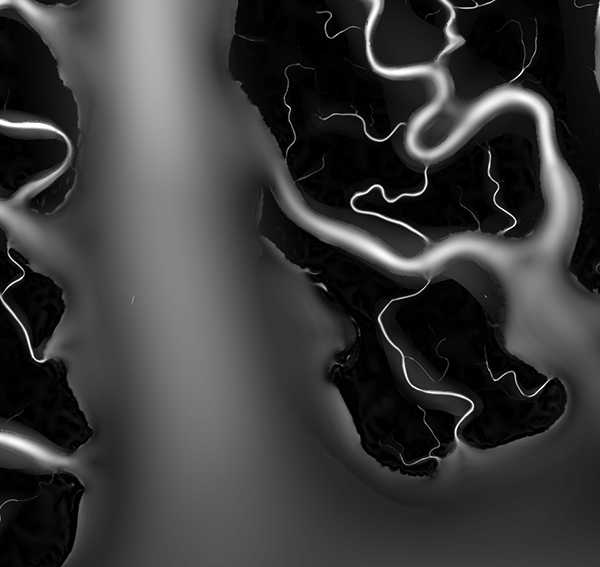}%
\label{mod_response_zoom}}
\caption{Comparison of original and modified Multiscale Singularity Index response. Full sized images: original response (a) and modified response (b). Zoomed images: original response (c) and modified response (d).}
\label{fig:mod-index}
\end{figure}

\noindent
The Multiscale Singularity Index creates ripples near the banks of wide rivers when a large range of scales is processed (Fig \ref{orig_response_zoom}). The ripples occur after finding the maximum response over all scales at each spatial coordinate. To attenuate the ripples, we employ an adaptive smoothing algorithm that adjusts the strength of smoothing based on the estimated scale for each pixel, so that coarse scales can be smoothed more than fine scales. To implement an adaptive smoothing algorithm in a computationally efficient way, we first compute an integral image over the Singularity Index response, which enables fast computation of summations over regions of arbitrary size. Then, we smooth the response using a box filter with a variable window size that is determined by the estimated scale. Since the integral image is computed only once, the algorithm only needs to perform two additions and one subtraction per pixel. We apply the adaptive box filter iteratively to approximate a Gaussian filter.

Typical results delivered by the Multiscale Singularity Index and our modified version are compared in Fig. \ref{fig:mod-index}.

\subsection{Centerline Extraction}
\noindent
To determine the channel centerlines, the maximum response across all scales is computed at each coordinate, and the orientation value at the maximum-response scale is taken to be the dominant centerline direction. A process of non-maxima suppression is applied along the dominant direction as explained in \cite{muralidhar2013steerable}. Then, a threshold level $T$ is determined on the non-maxima suppressed (NMS) image using Otsu's method \cite{otsu1975threshold}. To preserve channel connectivity, a hysteresis threshold is applied to binarize the NMS image, as follows:

\begin{enumerate}
\item Set pixels above an error threshold $\epsilon$ to one and the rest to zero
\item Find the connected components in the image
\item Find and remove the connected components that do not have at least one pixel above the threshold $T$.
\end{enumerate}

\begin{figure}[!t]
\centering
\includegraphics[width=3.1in]{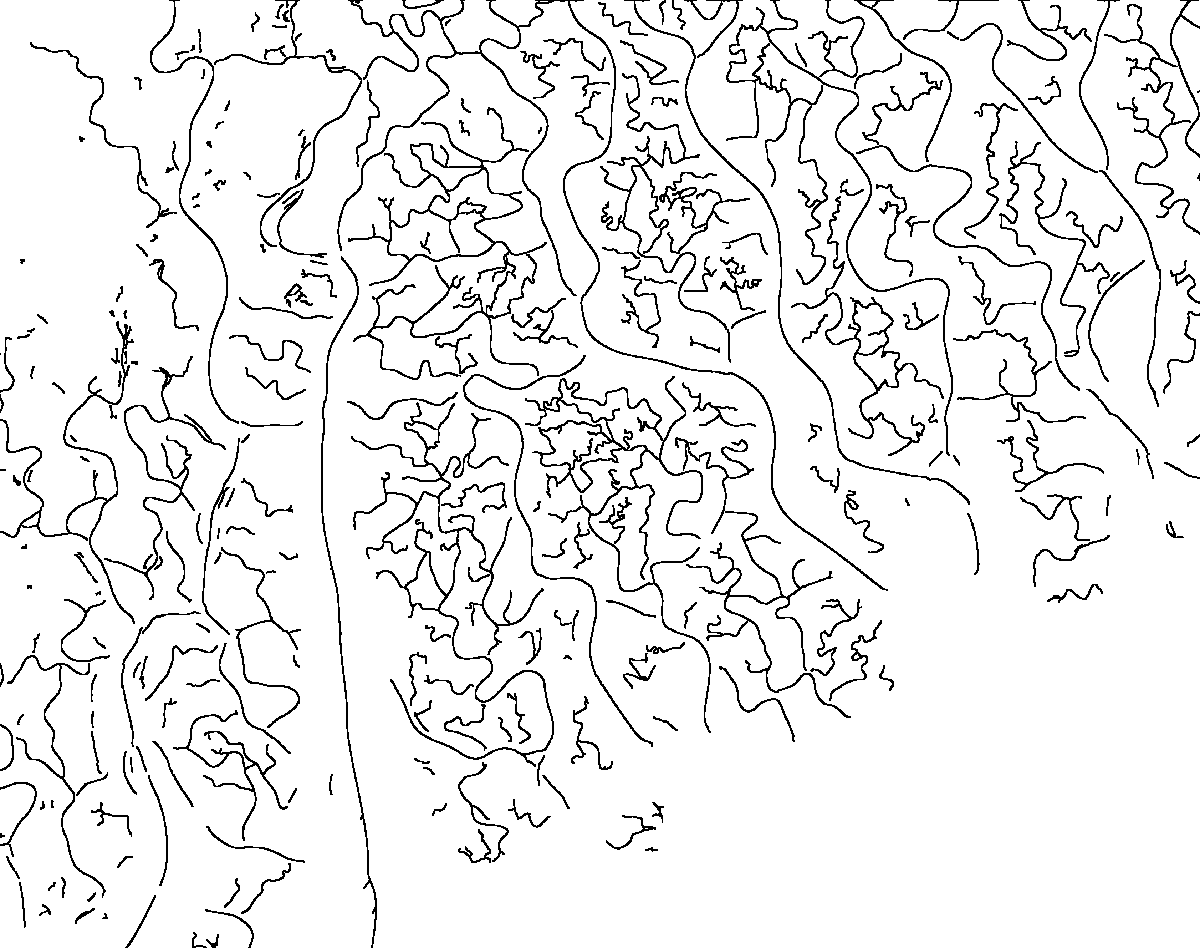}
\caption{Typical results of centerline extraction}
\label{fig:NMS}
\end{figure}

The error threshold $\epsilon$ is chosen empirically as $0.1 \times T$. Extracted centerlines for an example input image are illustrated in Fig. \ref{fig:NMS}. The figure is inverted for better visualization.

\subsection{Creating a Map of Channels}
\noindent
To show the computed channel width at each spatial coordinate along a centerline, a map of channels is created by re-growing the channels. The channels are re-grown by drawing a line of length $w(x,y)$ and orientation $\theta(x,y)$ at each spatial location $(x,y)$. The algorithm estimates the width of any water body of width smaller than the largest scale. Therefore, the resultant map includes ponds and other small water bodies as well as channels. Results of channel map creation are presented in Section \ref{sec:results}.

\section{Experiments and Results}
\label{sec:results}
\noindent
We tested our model on three different regions having different characteristics: Mississippi river near Memphis, TN (I1), Mississippi, Missouri, and Illinois rivers near St. Louis, MO (I2), and a portion of the Ganges-Brahmaputra-Jamuna Delta (I3). We used Landsat-8 images, downloaded at http://earthexplorer.usgs.gov/, to create the input images for our algorithm. The algorithm requires input images to have a contrast between water and non-water pixels. An example input could be a near-infrared image or a water index that uses multiple bands. In our experiments, the input images were created using the Modified Normalized Difference Water Index (MNDWI)\cite{xu2006modification}, which is an effective way to extract water information from remote sensing imagery.

The ground truth for I1 and I2 are obtained by aligning, cropping, and rasterizing the river data from the National Hydrography Dataset (http://nhd.usgs.gov/). For I3, the Ganges-Brahmaputra-Jamuna (GBJ) Delta network extracted by \cite{passalacqua2013geomorphic} is used as the ground truth. The extraction performed by \cite{passalacqua2013geomorphic} included manual cleaning and a comparison to Google Earth imagery.

We fixed the minimum scale $\sigma_1$ to its default value \cite{muralidhar2013steerable}, 1.5 pixels, which is the smallest width for a channel to be captured by the algorithm. The number of scales is determined automatically using the upper bound that is described earlier. To reduce the computation time, a smaller number of scales could also be chosen if all channels of interest are known to be smaller than a certain width. In the experiments reported here, we set the number of scales $N$ to 16.

\begin{figure*}[!t]
\centering
\subfloat[]{
\includegraphics[height=2.1in]{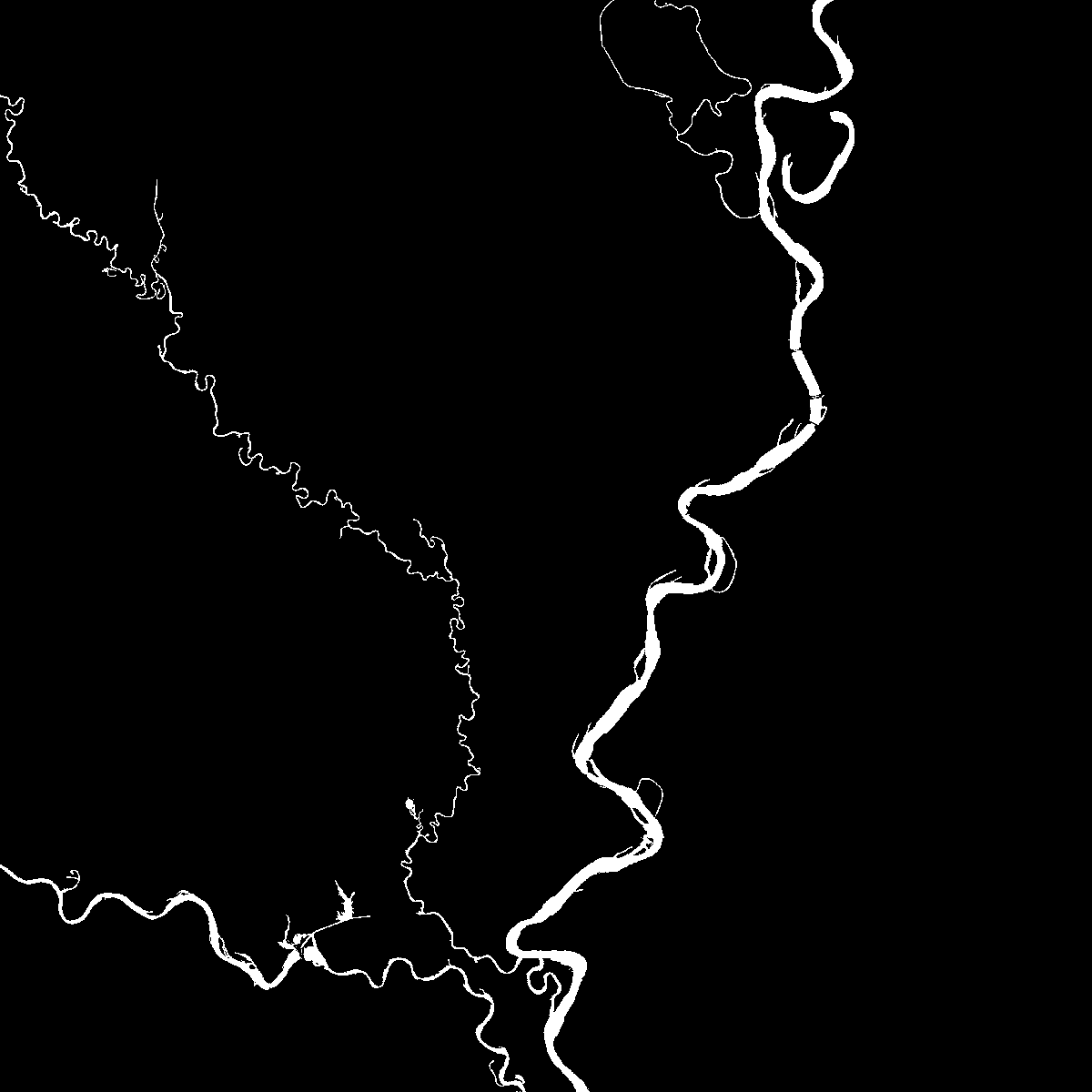}%
~
\includegraphics[height=2.1in]{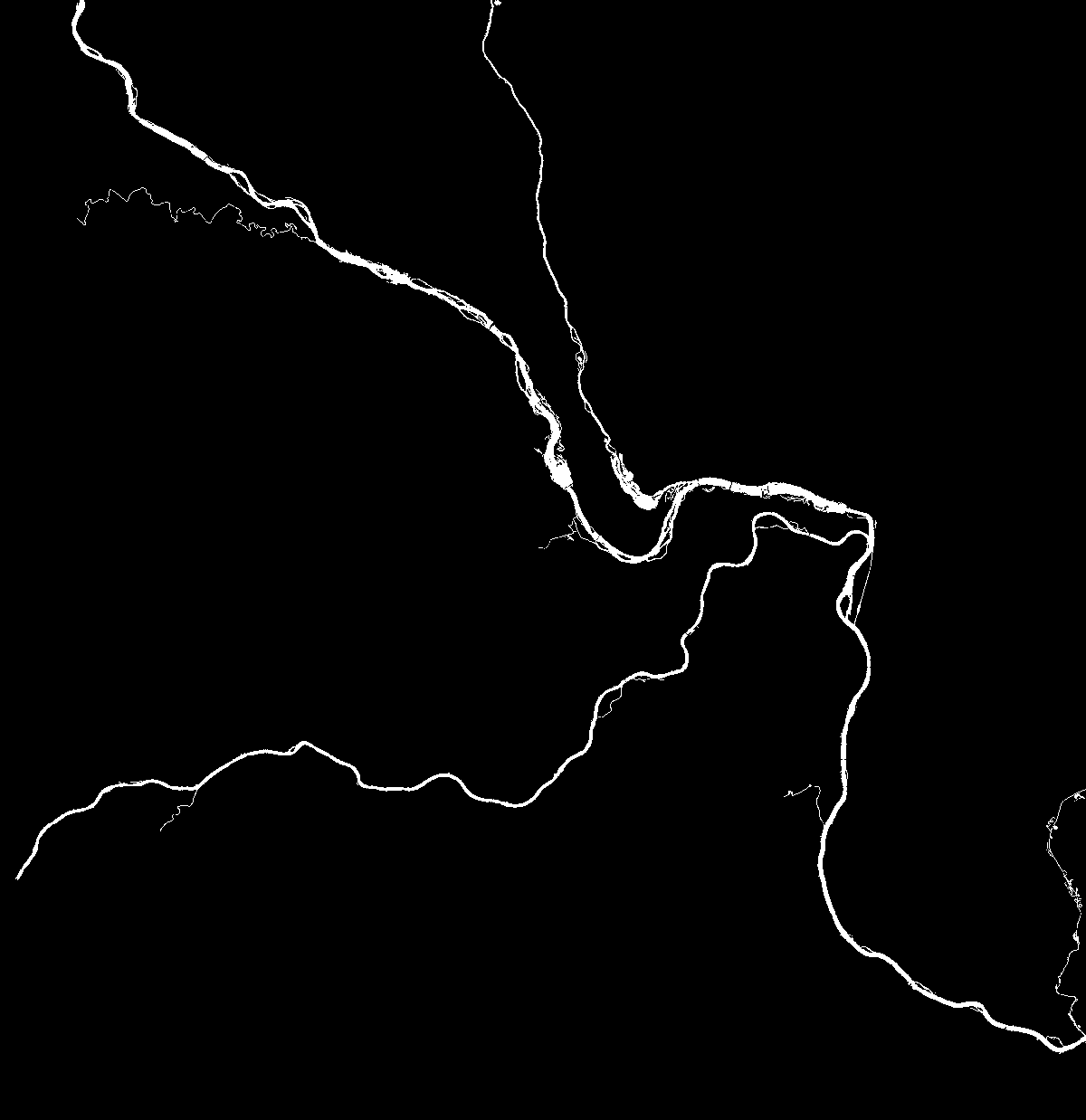}%
~
\includegraphics[height=2.1in]{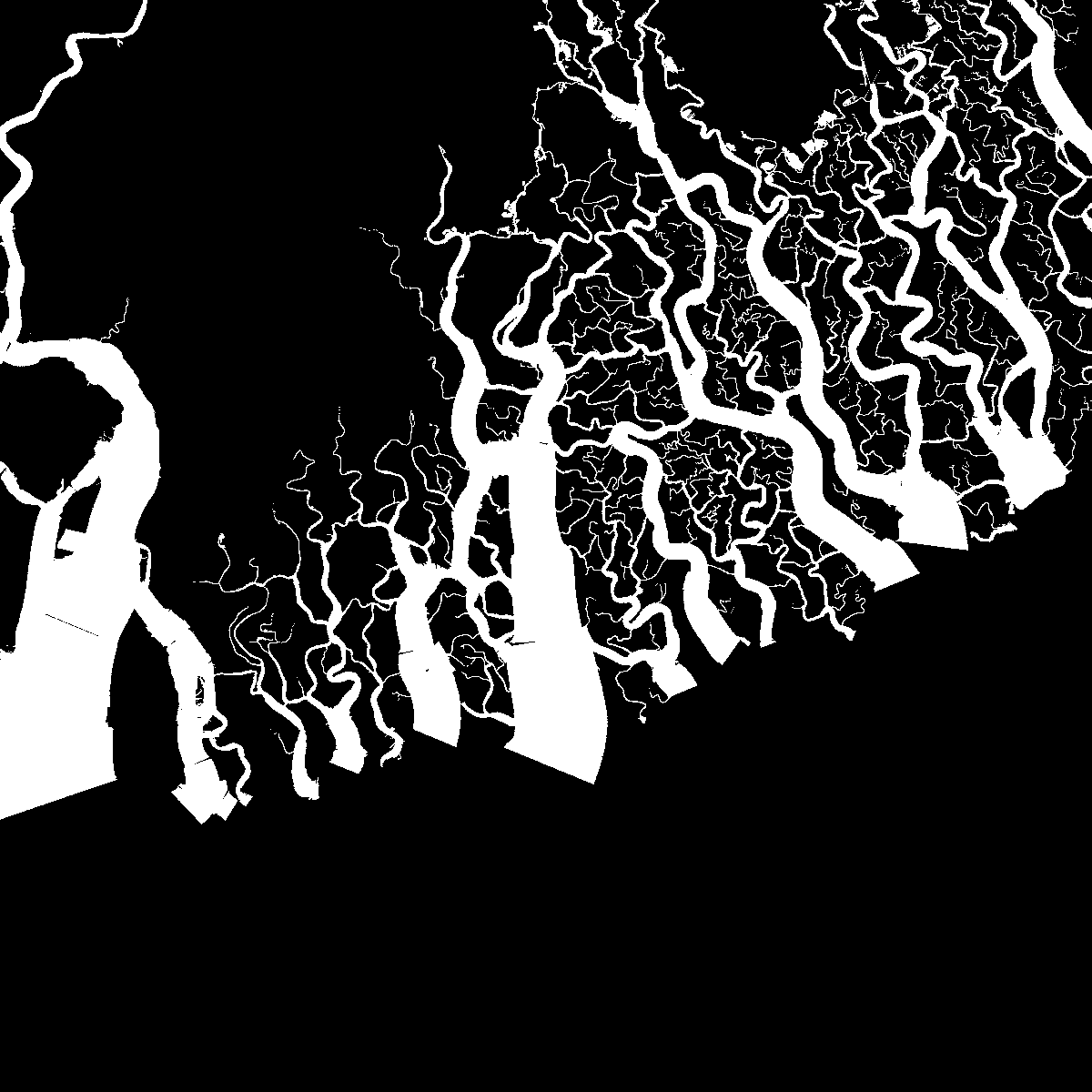}%
}

\subfloat[]{
\includegraphics[height=2.1in]{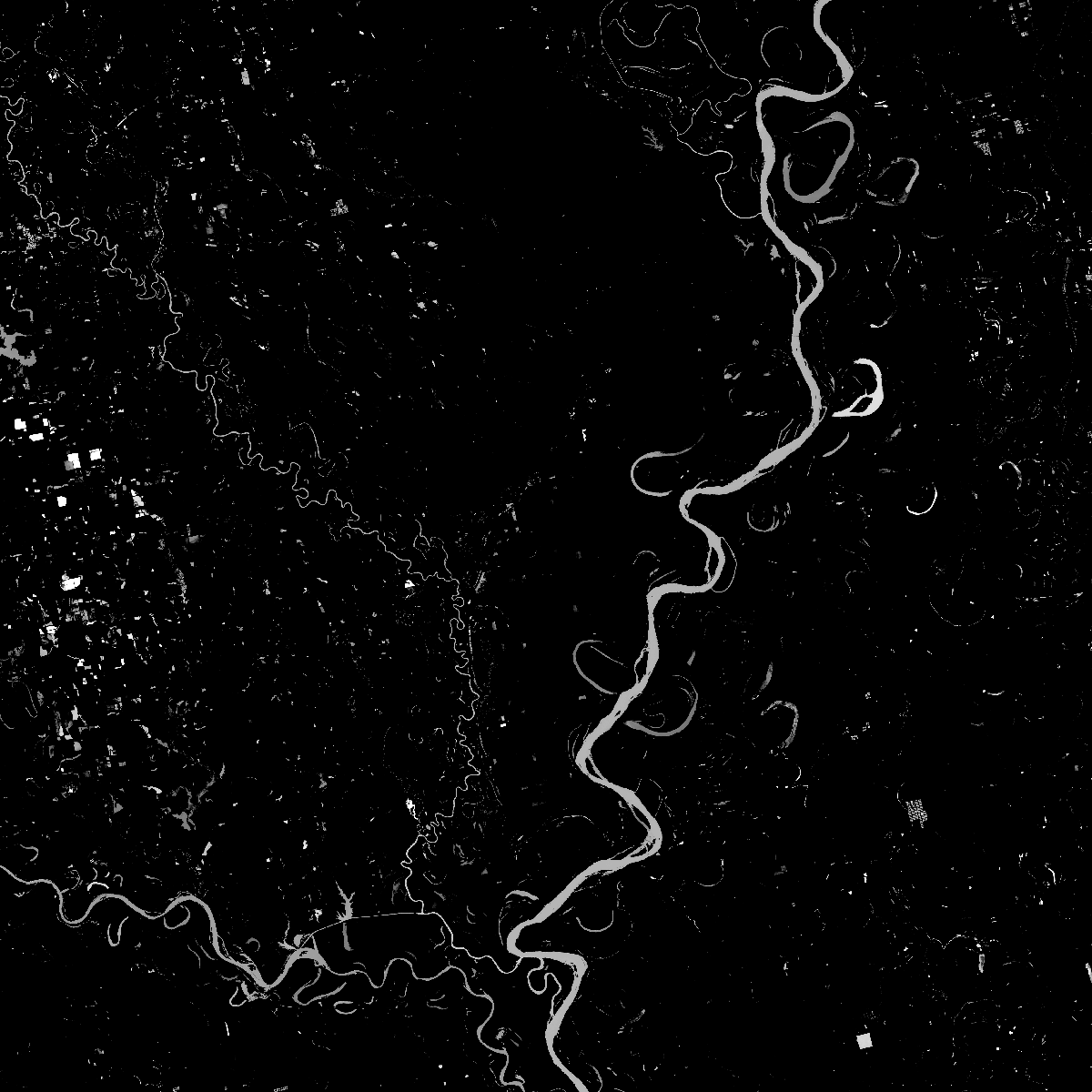}%
~
\includegraphics[height=2.1in]{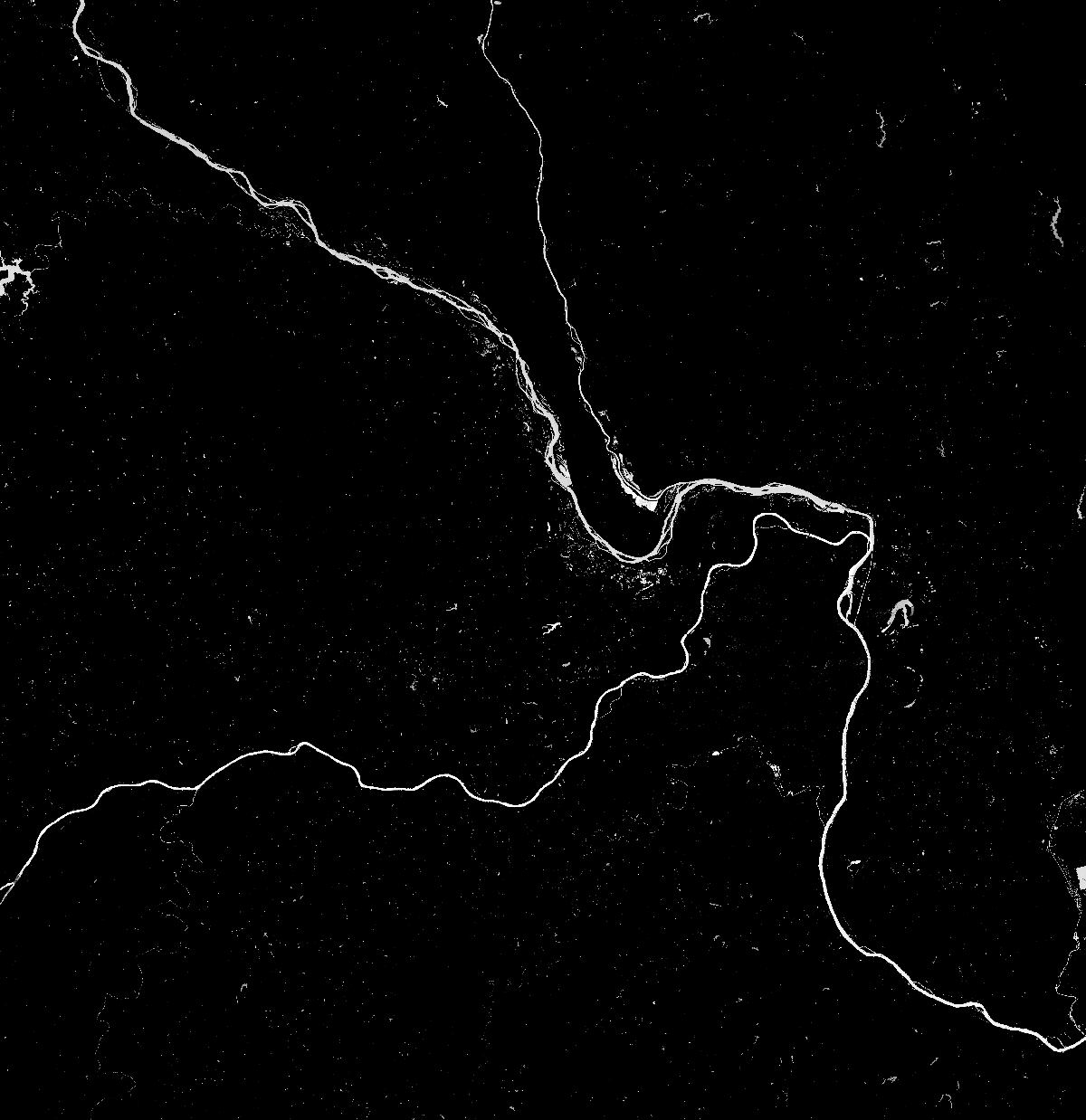}%
~
\includegraphics[height=2.1in]{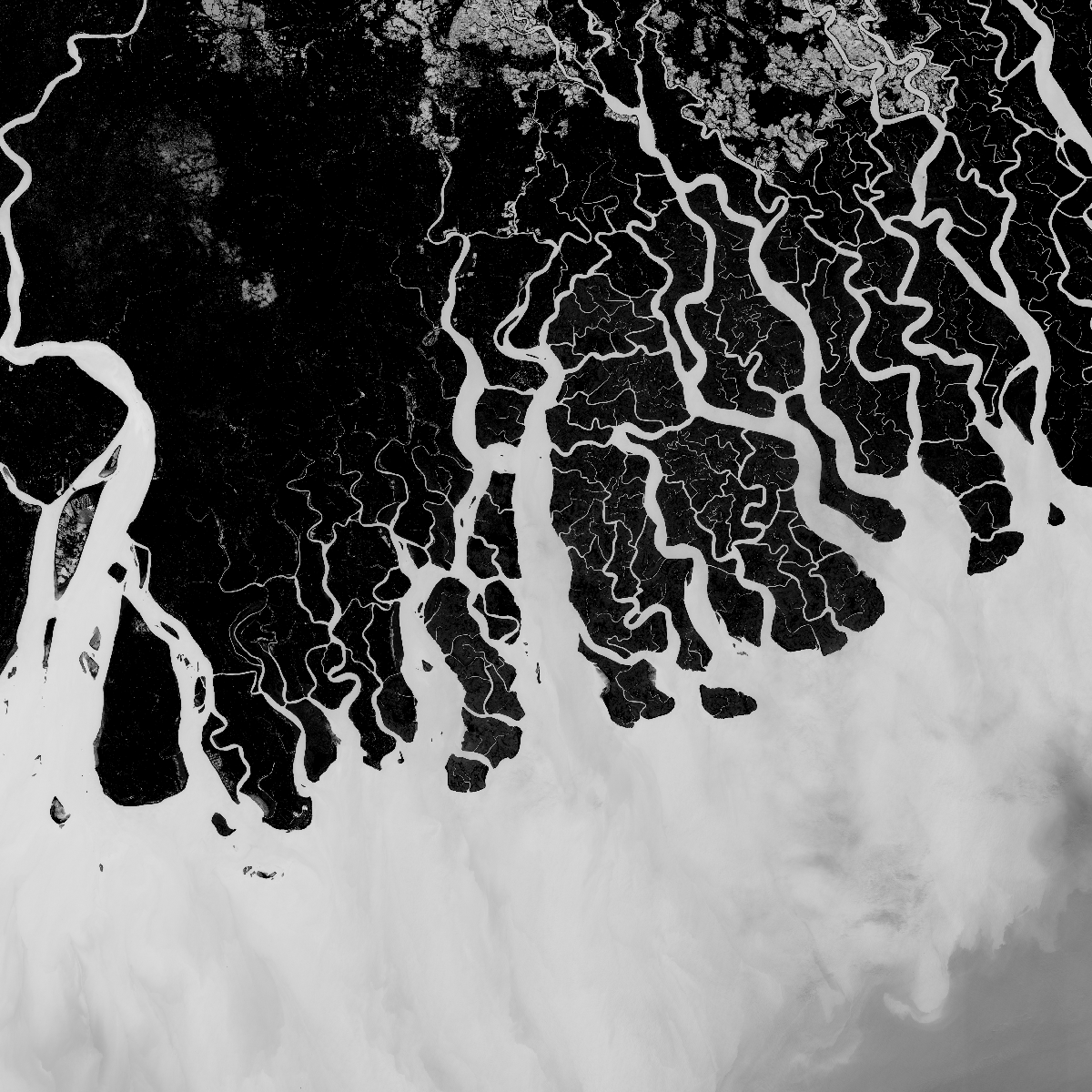}%
}

\subfloat[]{
\includegraphics[height=2.1in]{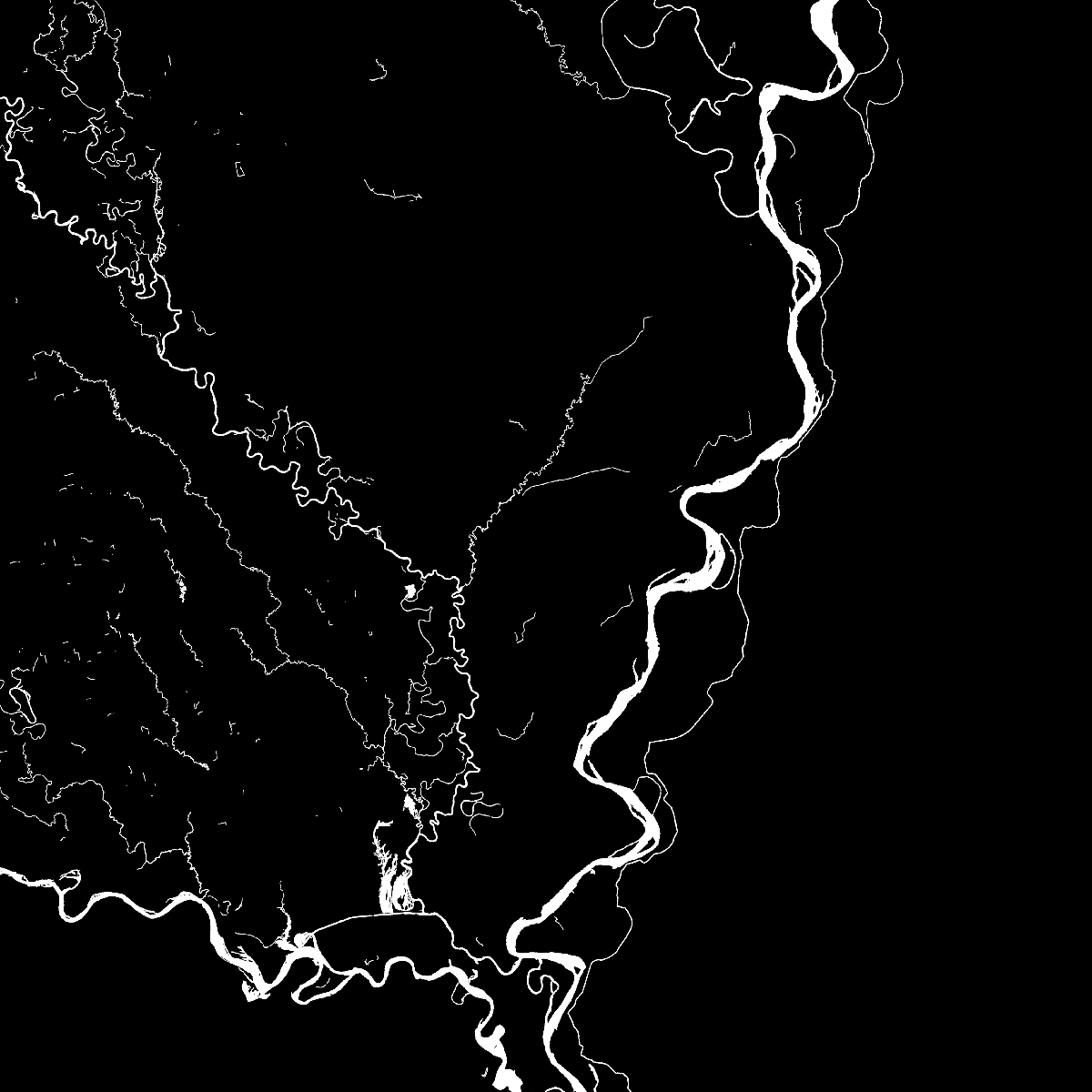}%
~
\includegraphics[height=2.1in]{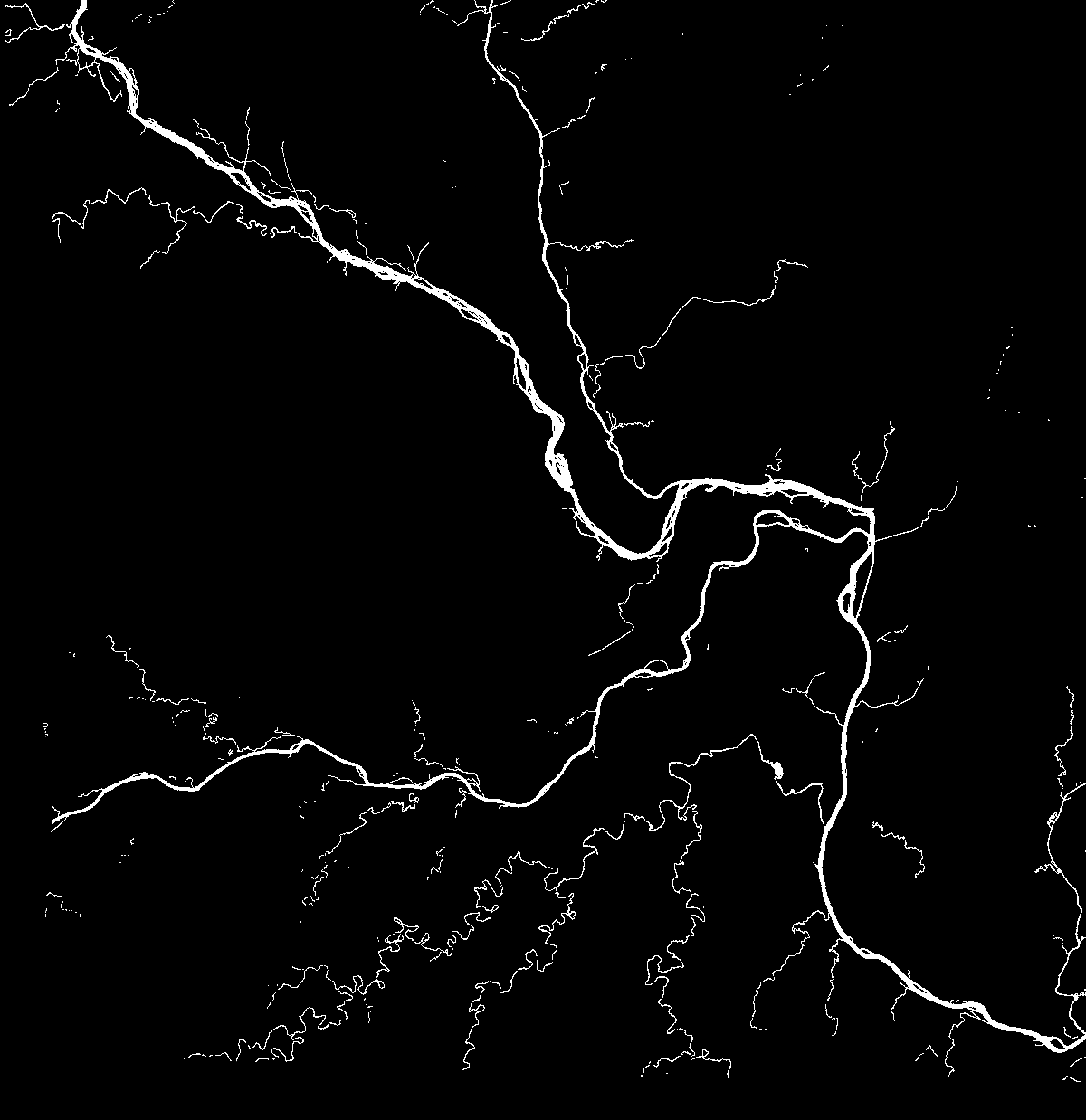}%
~
\includegraphics[height=2.1in]{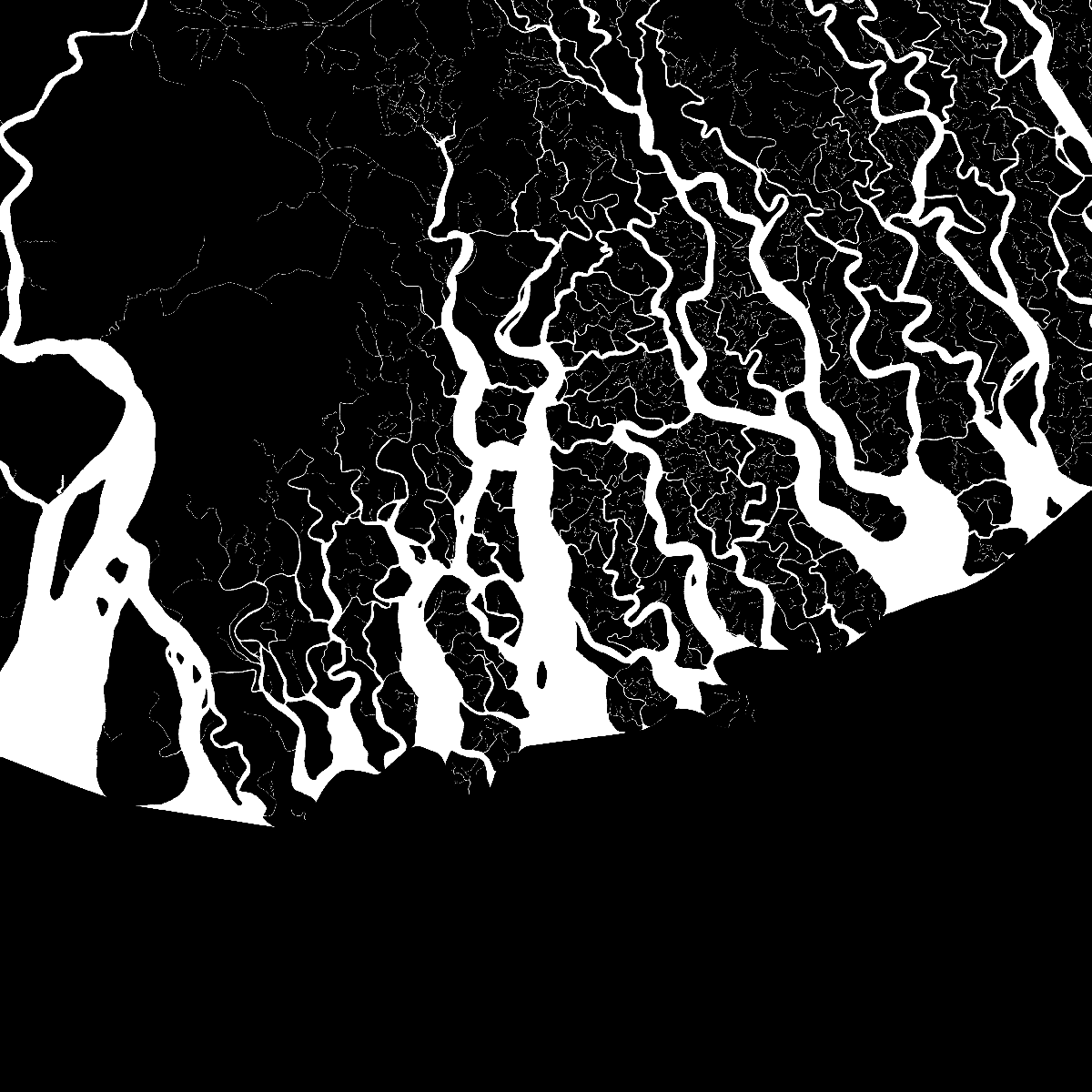}%
}
\caption{Comparison of the regrown channel maps (a) with the input (b) and ground truth (c) images. First column: Mississippi river near Memphis, TN; second column: Mississippi, Missouri, and Illinois rivers near St. Louis, MO; and third column: a portion of the Ganges-Brahmaputra-Jamuna Delta. Landsat IDs of the scenes: LC80230362015045LGN00, LC80240332014305LGN00, and LC81380452014304LGN00.}
\label{fig:results}
\end{figure*}

The regrown channel maps, showing the estimated location, width, and orientation of the channels, are compared with the ground truth and input images in Fig. \ref{fig:results}. The ground truth images did not include non-channel water bodies. Therefore, we removed the non-channel water bodies also from the regrown channel maps, by discarding the connected components that constitute less than 0.1\% of the maps. Given the ground truth images, the accuracies ($(TP + TN)/(TP + TN + FP + FN)$) of the re-grown channel network images were found to be 96.77\%, 97.86\%, and 91.13\% for I1, I2, and I3, respectively. 

\section{Conclusion and Future Work}
\noindent
We have described an automatic channel network extraction algorithm that inputs remotely sensed images and produces maps of estimated channel centerline, width, and orientation. We modified a Multiscale Singularity Index to extract a network of channels over a wide range of scales. The algorithm works automatically without any user intervention.

Our method can be used to analyze channel networks in different environments and over time to capture the effect of environmental forcing and natural and anthropogenic change on the network structure. One of our future research directions is to analyze deltaic response to anthropogenic and natural forcings in coastal areas. We also plan to extend our work towards automatically creating topological maps, which will provide graph representations of channel networks.

\section*{Acknowledgment}
\noindent
The authors would like to acknowledge support from the National Science Foundation, grant numbers CAREER/EAR--1350336 and FESD/EAR--1135427 awarded to P.P.

\ifCLASSOPTIONcaptionsoff
  \newpage
\fi



%



\bibliographystyle{IEEEtran}
\bibliography{mybib}{}

%








\end{document}